%% file: root.tex
\documentclass[letterpaper, 10pt, conference]{ieeeconf}

\IEEEoverridecommandlockouts                              

\title{\LARGE \bf
Fast Fault Detection on a Quadrotor using Onboard Sensors and a Kalman Filter Approach
}

\author{B.A. Strack van Schijndel$^{1}$, S. Sun$^{2}$ and C.C. de Visser$^{3}$

\thanks{$^{1}$Graduate student, Section Control and Simulation, Faculty of Aerospace Engineering, Delft University of Technology, The Netherlands {\tt\small bramsvs@gmail.com}}%
\thanks{$^{2}$Ph.D. student, Section Control and Simulation, Faculty of Aerospace Engineering, Delft University of Technology, The Netherlands {\tt\small s.sun-4@tudelft.nl}}%
\thanks{$^{3}$Assistant professor, Section Control and Simulation, Faculty of Aerospace Engineering, Delft University of Technology, The Netherlands {\tt\small c.c.devisser@tudelft.nl}}%
}

\usepackage{tikz}
\usetikzlibrary{shapes,arrows}

\usepackage{multirow}
\usepackage{hyperref}
\usepackage{amsmath}
\usepackage{graphicx}
\usepackage{import}
\usepackage{newtxmath}

\usepackage{siunitx}
\usepackage{nomencl}
\makenomenclature

\usepackage{etoolbox}
\usepackage{booktabs}
\usepackage{multicol}

\newcommand{\nomunit}[1]{%
\renewcommand{\nomentryend}{\hspace*{\fill}[#1]}}

\newcommand{\w}[1]{\omega_{#1}^2}

\usepackage{algorithm2e}
\SetKwInOut{Input}{Input}
\SetKwInOut{Output}{Output\,}
\SetAlgoLined
\DontPrintSemicolon

\newcommand{\vect}[1]{\boldsymbol{#1}}

\newcommand{\tx}{\text}

\newcommand{\kthres}{k_\tx{thres}}
\newcommand{\pthres}{(P_\tx{fail})_\tx{thres}}

\newcommand{\kiest}{\hat{k}_i}
\newcommand{\var}{\text{Var}}

\begin{document}

\maketitle
\thispagestyle{empty}
\pagestyle{empty}

\begin{abstract}
  This paper presents a novel method for fast and robust detection of actuator failures on quadrotors. The proposed algorithm has very little model dependency. 
  A Kalman estimator estimates a stochastic effectiveness factor for every actuator, using only onboard RPM, gyro and accelerometer measurements. 
  Then, a hypothesis test identifies the failed actuator. 
  This algorithm is validated online in real-time, also as part of an active fault tolerant control system. Loss of actuator effectiveness is induced by ejecting the propellers from the motors.
  The robustness of this algorithm is further investigated offline over a range of parameter settings by replaying real flight data containing 26 propeller ejections.
  The detection delays are found to be in the 30$\sim$130 ms range, without missed detections or false alarms occurring.
\end{abstract}

\nomenclature[B]{$(\cdot)_\tx{sp}$}{Setpoint}
\nomenclature[B]{$(\cdot)_k$}{Timestep}
\nomenclature[B]{$(\cdot)_f$}{Filtered}
\nomenclature[B]{$(\cdot)_i$}{Actuator index}
\nomenclature[B]{$(\cdot)_b$}{Body axis}

\nomenclature[D]{$\vect{K}$}{Kalman gain vector \nomunit{-}}
\nomenclature[D]{$\vect{R}$}{Measurement noise matrix}
\nomenclature[D]{$\vect{Q}$}{Process noise matrix}
\nomenclature[D]{$\vect{P}$}{Covariance matrix}
\nomenclature[D]{$\vect{H}$}{Observation matrix}
\nomenclature[D]{$H$}{Laplace transform}

\nomenclature[D]{$\vect{x}$}{State vector}
\nomenclature[D]{$\vect{y}$}{Innovation vector}
\nomenclature[D]{$\vect{\Omega}$}{Vehicle angular rate vector \nomunit{\si{\radian\per\second}}}

\nomenclature[D]{$\vect{I}_v$}{Identity/inertia matrix}
\nomenclature[D]{$b$}{Actuator x-position \nomunit{\si{\metre}}}
\nomenclature[D]{$h$}{Actuator y-position \nomunit{\si{\metre}}}
\nomenclature[D]{$\vect{d}$}{Disturbance vector \nomunit{-}}
\nomenclature[D]{$\omega$}{Angular rate \nomunit{\si{\radian\per\s}}}
\nomenclature[D]{$\tau$}{Torque \nomunit{\si{\newton\meter}}}
\nomenclature[D]{$I$}{Vehicle inertia \nomunit{\si{\kilogram\per\metre\squared}}}
\nomenclature[D]{$m$}{Vehicle mass \nomunit{\si{\kilogram}}}
\nomenclature[D]{$p$}{Roll rate \nomunit{\si{\radian\per\second}}}
\nomenclature[D]{$q$}{Pitch rate \nomunit{\si{\radian\per\second}}}
\nomenclature[D]{$\vect{r}$}{Actuator arm vector \nomunit{\si{\metre}}}
\nomenclature[D]{$c_T$}{Propeller thrust coefficient \nomunit{\si{\N\s\squared}}}
\nomenclature[D]{$c_M$}{Propeller moment coefficient \nomunit{\si{\N\metre\s\squared}}}
\nomenclature[D]{$k_i$}{Actuator effectiveness scaling factor of actuator $i$ \nomunit{-}}
\nomenclature[D]{$\sigma^2_{k_i}$}{Variance of $k_i$ \nomunit{-}}
\nomenclature[D]{$P$}{Probability \nomunit{-}}

\nomenclature[D]{$\Delta t$}{Step size \nomunit{\si{\second}}}

\nomenclature[D]{$\omega_n$}{Natural frequency \nomunit{\si{\per\second}}}
\nomenclature[D]{$\zeta$}{Damping ratio \nomunit{-}}
\nomenclature[D]{$s$}{Laplace variable \nomunit{-}}


\section{Introduction}



Loss of actuator effectiveness (LOE) is one of the many \textit{system failures}~\cite{belcastroExperimentalFlightTesting2017} that could happen.
LOE can happen suddenly due to, for example, propeller faults or other structural failures such as motor arm or assembly breakage. 
Quadrotors especially lack redundancy in their actuators making actuator faults risky. One way to cope with (single) actuator failures on quadrotors is to sacrifice yaw control and use the remaining rotors to land directly~\cite{muellerStabilityControlQuadrocopter2014}, or maintain forward flight~\cite{sunHighSpeedFlightQuadrotor2018}.
In order to apply an appropriate control strategy, these active fault tolerant control methods require a quick loss of effectiveness detection. 
Also without sacrificing overall system reliability.


More generally, a fault tolerant control system should prevent simple failures from developing into catastrophic failures. Instead, the system should gracefully degrade by giving up certain functions while maintaining (some) control over other. This is a desirable property of a quadrotor because it improves its safety. 




The detection scheme presented in this paper takes RPM, gyro and accelerometer measurements as inputs and dependents on a very limited control effectiveness model. A Kalman estimator estimates a probabilistic LOE-factor for each actuator. Then, a hypothesis test identifies the faulty actuator.
In order to arrive at the limited model assumptions had to be made. In order to test these assumptions flight tests were done on a Parrot Bebop 2 quadrotor unmanned air vehicle (UAV), validating the detection scheme.

In summary, the contributions of this research is:

\begin{enumerate}
  \item Split the actuator fault detection problem in an actuator dynamics part and an actuator effectiveness part (as in INDI control~\cite{smeurAdaptiveIncrementalNonlinear}), by utilizing RPM feedback measurements. Then provide a solution to the loss of actuator effectiveness detection problem, by first applying a Kalman estimator, then a hypothesis test. (Section~\ref{sec:method})
  \item Validate its working in both real-time online and replayed offline environments, at scale across many different flights and varying parameters. (Section~\ref{sec:results})
\end{enumerate}

\section{Related work}

Most work on LOE-detection for quadrotors has been on partial (10\%-60\%) LOE. Recently Nguyen and Hong proposed a method based on a sliding mode Thau observer~\cite{nguyenSlidingModeThau2018}. Zhong, Zhang et al. proposed a three-stage Kalman filter approach to deal with external disturbances \cite{zhongRobustActuatorFault2018}. Avram et al. systematically designed and flight tested a method based on a fault detection and isolation framework \cite{avramQuadrotorActuatorFault2017a}.
Lu and Van Kampen estimated LOE-factors by taking the pseudoinverse of the effectiveness matrix \cite{luActiveFaulttolerantControl2015}, this method is unfortunately very sensitivity noise.
Some methods require model information such as quadrotor inertia or mass~\cite{freddiActuatorFaultDetection2010, luActiveFaulttolerantControl2015, avramQuadrotorActuatorFault2017, zhongRobustActuatorFault2018,hasanModelBasedActuatorFault2018}, knowledge that is often not readily available in real-world scenarios. Also the availability of inputs such as attitude~\cite{freddiActuatorFaultDetection2010, luActiveFaulttolerantControl2015, zhongRobustActuatorFault2018} or position~\cite{freddiActuatorFaultDetection2010, zhongRobustActuatorFault2018} is often assumed.

Regarding other types of multirotor UAV, Frangenberg et al. developed and flight tested a method based on a bank of WLS-estimators on a octorotor~\cite{frangenbergFastActuatorFault2015}. Vey and Lunze investigated an approach based on a Luenberger observer on a hexarotor UAV~\cite{veyExperimentalEvaluationActive2016}.


There are techniques for detecting other classes of actuator faults that not \textit{necessarily} result in loss of effectiveness.
Vibration based methods exploit propeller or motor imbalances, challenging is the isolation to a specific actuator. Jiang et al. \cite{jiangFaultDetectionIdentification2015} propose a method that includes feature extraction using wavelets. These features then serve as an input to train an artificial neural network, after which this network can detect fractured and distorted propellers.
Ghalamchi and Mueller have taken steps to detect \textit{and isolate} faults by applying a Fourier transform \cite{ghalamchiVibrationBasedPropellerFault2018}. Future work is to do the fault detection automatically in real-time.
Other authors exploit motor current and/or acoustics measurements \cite{brownCharacterizationPrognosisMultirotor2015, misraStructuralHealthMonitoring2018}.

Unrelated to UAVs, Wu et al.~\cite{wuControlEffectivenessEstimation1998} applied a Kalman estimator for estimating the changes in control effectiveness of the control surfaces of civil aircraft.


\section{Problem formulation} \label{sec:problem}

The aim of this work is to detect loss of actuator effectiveness, by using onboard sensor measurements only, and validate the method on a real quadrotor in real world conditions. As a starting point a fault scenario is assumed.


\subsection{Fault scenario}
An actuator fault often does not happen spontaneously. It could be the result of an impact from another flying, static or moving object. One could think of an in-flight collision with another UAV. 

In this research, we do not simulate an actuator failure by corrupting the rotor speed setpoint $\omega_\text{sp}$, but instead we induce a real actuator failure by ejecting the propeller in-flight. This has the following two important implications: 
1) the RPM-measurements \textit{resulting} from the rotor speed setpoint are also \textit{not} corrupted, thus can be used as input to the detection algorithm, and 
2) the loss of effectiveness of an actuator is sudden and total, so no spin-down lag or percentage reduction of effectiveness.

For these reasons the propeller ejection method is critical to the detection method presented in this research.


\subsection{Propeller ejection method} \label{subs:eject_method}

\begin{figure}[thpb]
  \centering
  \includegraphics[scale=0.035]{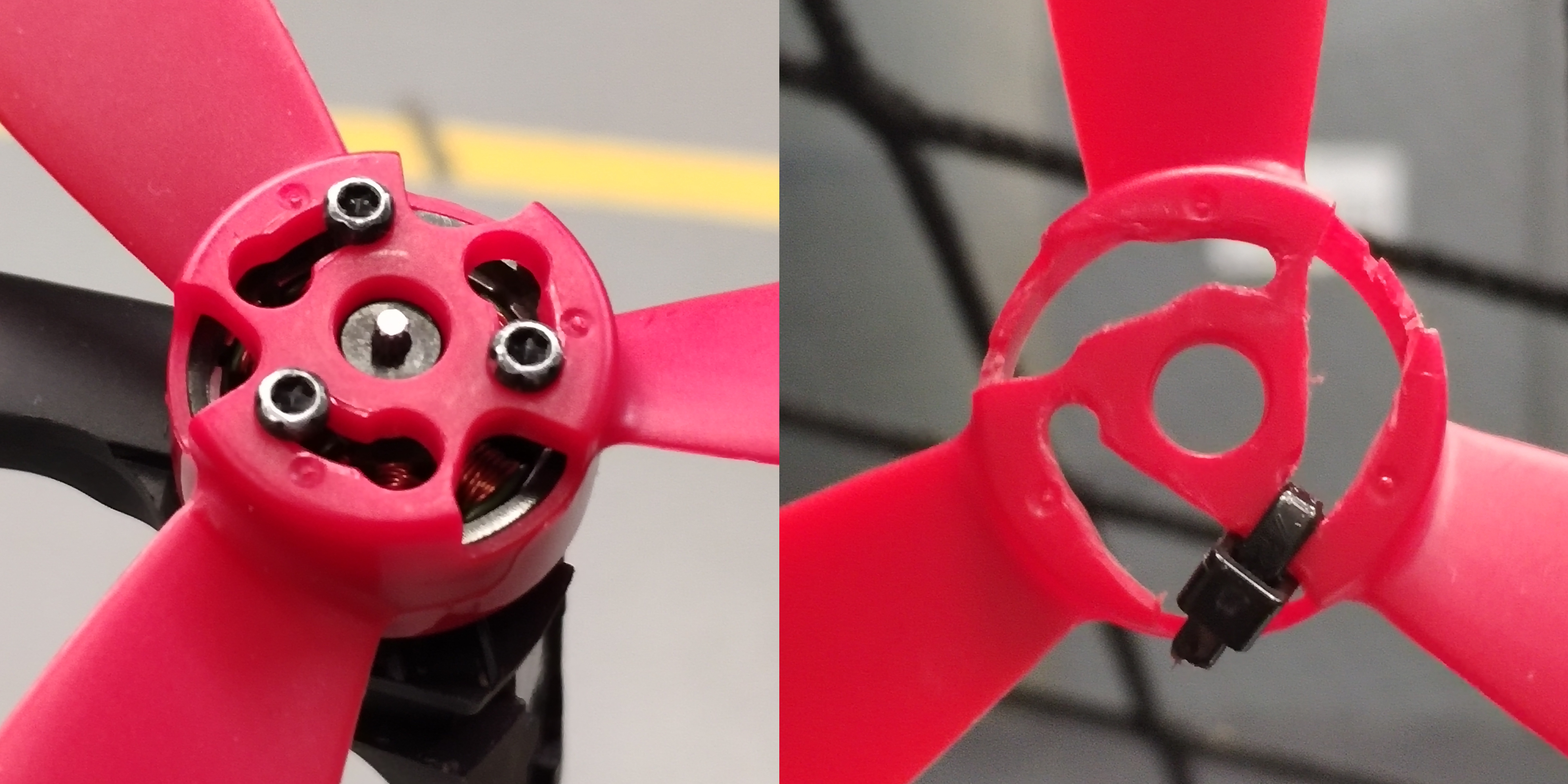}
  \caption{Nominal locking mechanism (left),  frictionless mechanism (right).}
  \label{fig:props}
\end{figure}

The propeller ejection method was developed specifically for the Bebop 2, yet it can be generalized.
The main idea is to remove the friction in the locking mechanism (Figure~\ref{fig:props}). Then keep the propeller attached using the aerodynamic forces, and eject it by quickly reducing the motor speed significantly. The assumption is that if the rotational deceleration of the propeller only subject to the aerodynamic torque:
\begin{equation*}
  (\dot{\omega}_\text{prop})_\text{aero} = \ 
  \dfrac{(\tau_\text{prop})_\text{aero}}{I_\text{prop}}
\end{equation*}
is less than the deceleration of the motor:

\begin{equation*}
  \dot{\omega}_\text{motor} < (\dot{\omega}_\text{prop})_\text{aero}
\end{equation*}
the propeller will eject.

The motor speed can be reduced sufficiently fast to cause an ejection using one of the following three methods: 1) by setting a lower altitude setpoint, 2) by giving yaw rate commands, or 3) by adding a sawtooth signal to the actuator setpoint $\omega_\text{sp}$. On the contrary, an ejection can be prevented by artificially lowering the control gains.

\begin{figure}
  \centering
  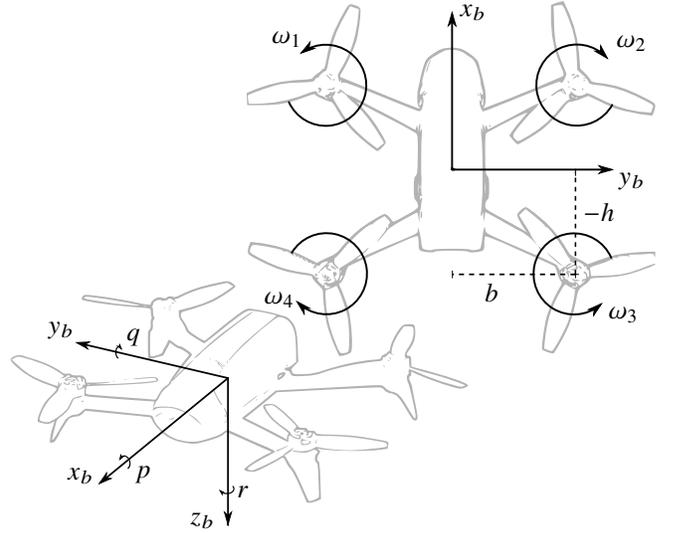
  \caption{Geometry of the Parrot Bebop 2 quadrotor with rotor speeds $\omega_i$, geometry $b$ and $h$, body axes $(\cdot)_b$ and angular rates $p, q, r$.}
  \label{fig:bebop2}
\end{figure}

\subsection{Control effectiveness model}
All quantities are projected onto the body frame. Figure~\ref{fig:bebop2} defines its axes, the geometry $b, h$ and the angular velocities of the actuators $\omega$.
Assuming in flight,~\eqref{eq:euler_moments} and~\eqref{eq:newton_accelerations} describe the dynamics of the quadrotor:
\begin{equation} \label{eq:euler_moments}
  \vect{M}_\text{act} + \vect{M}_\text{aero} + \vect{M}_\text{gyro} = \vect{I}_v \vect{\dot{\Omega}} + \vect{\Omega} \times \vect{I}_v \vect{\Omega}
\end{equation}
with the summed actuator moments $\vect{M}_\text{act}$, aerodynamic moment $\vect{M}_\text{aero}$, gyroscopic moments $\vect{M}_\text{gyro}$, inertia matrix $\vect{I}_v$ and angular rate vector $\vect{\Omega}=\begin{bmatrix}p & q & r \end{bmatrix}$. 
  
Also assuming in flight, the linear acceleration in z-direction can be modeled as:
\begin{equation} \label{eq:newton_accelerations}
  - \sum^4_{i=1} (T_\text{act})_i + (F_\text{aero})_z = m \cdot a_z
\end{equation}
with actuator thrust force $T_\text{act}$, aerodynamic force $F_\text{aero}$, quadrotor mass $m$ and proper acceleration $a_z$. That acceleration is measured by the accelerometer. Considering accelerometer measurements, only the measurements in $z$-direction are of interest assuming the actuator thrust points in that direction. 


The actuator thrust forces are assumed to be linear with respect to the square of the rotor speed:
\begin{equation} \label{eq:thrust}
  (T_\text{act})_i = c_T k_i \omega_i^2
\end{equation}
with actuator thrust coefficient $c_T$ and actuator effectiveness scaling factors $k_i$. A value of $k_i = 1$ corresponds to nominal effectiveness while $k_i = 0$ corresponds to complete loss of effectiveness. These scaling factors $k_i$ will be estimated within the LOE-detection algorithm as explained in Section~\ref{sec:method}.




The moment generated by the actuator thrust forces can be modeled as:
\begin{equation} \label{eq:thrust_moments}
  \vect{M}_{\text{act}} =
  \sum^4_{i=1} \left( (\vect{r}_{\text{act}})_i \times
  \begin{bmatrix}
    0 \\ 0 \\ -(T_\text{act})_i
  \end{bmatrix}
  \right)
\end{equation}
with position vectors of the actuators $(r_\text{act})_i$:
\begin{equation*}
  \vect{r}_1 = 
  \begin{bmatrix}
    h \\ -b \\ 0
  \end{bmatrix},\;
  \vect{r}_2 = 
  \begin{bmatrix}
    h \\ b \\ 0
  \end{bmatrix},\;
  \vect{r}_3 = 
  \begin{bmatrix}
    -h \\ b \\ 0
  \end{bmatrix},\;
  \vect{r}_4 = 
  \begin{bmatrix}
    -h \\ -b \\ 0
  \end{bmatrix}
\end{equation*}

Solving~\eqref{eq:euler_moments} for rotational acceleration gives:
\begin{equation} \label{eq:rot_acc}
  \vect{\dot{\Omega}} = \vect{I}_v^{-1}(\vect{\Omega} \times \vect{I}_v \vect{\Omega} - \vect{M}_\text{act} - \vect{M}_\text{aero} - \vect{M}_\text{gyro})
\end{equation}
This model can be simplified to:
\begin{equation} \label{eq:euler_moments_simple}
  \vect{\dot{\Omega}} = \vect{I}_v^{-1}(\vect{M}_\text{act}) + \vect{d}
\end{equation}
with disturbance $\vect{d}$ containing $\vect{M}_\text{aero}$, $\vect{M}_\text{gyro}$ the coupling term $\vect{\Omega} \times \vect{I}_v \vect{\Omega}$ and other phenomena such as propeller imbalances, body deformations and center of gravity offsets.
We assume that under normal flight conditions $\vect{d} \ll \vect{I}_v^{-1}(\vect{M}_\text{act})$.

Likewise,~\eqref{eq:newton_accelerations} can be solved for $a_z$ and simplified to:
\begin{equation} \label{eq:newton_accelerations_simple}
  a_z = m^{-1} \sum^4_{i=1} (T_\text{act})_i + d
\end{equation}

By assuming a diagonal vehicle inertia matrix $\vect{I}_v$, substituting  \eqref{eq:thrust_moments} in \eqref{eq:euler_moments_simple}, \eqref{eq:thrust} in \eqref{eq:newton_accelerations_simple} and working out the result in a single matrix equation, the rotational and linear accelerations can be modeled as:
\begin{equation}
  \begin{bmatrix}
    \dot{p} \\
    \dot{q} \\
    a_z     \\
  \end{bmatrix}
  =
  c_T
  \text{diag}\left(
  \begin{bmatrix}
    I_x \\
    I_y \\
    m   \\
  \end{bmatrix}
  \right)^{-1}
  \begin{bmatrix}
    b  & -b & -b & b  \\
    h  & h  & -h & -h \\
    -1 & -1 & -1 & -1 \\
  \end{bmatrix}
  \text{diag}(\vect{k})
  \begin{bmatrix}
    {\omega_1^2} \\
    {\omega_2^2} \\
    {\omega_3^2} \\
    {\omega_4^2} \\
  \end{bmatrix}
  + \vect{d}
\end{equation}
with $\vect{k} = \begin{bmatrix} k_1 & k_2 & k_3 & k_4\end{bmatrix}$,
which by defining:
\begin{equation*}
  G_p     := \dfrac{c_T b}{I_x},\;
  G_q     := \dfrac{c_T h}{I_y},\;
  G_{a_z} := \dfrac{c_T}{m}
\end{equation*}
can be rewritten as:
\begin{align} \label{eq:control_eff}
  \begin{bmatrix}
    \dot{p} \\
    \dot{q} \\
    a_z     \\
  \end{bmatrix}
   & =
  \text{diag}(\vect{G})
  \begin{bmatrix}
    1  & -1 & -1 & 1  \\
    1  & 1  & -1 & -1 \\
    -1 & -1 & -1 & -1 \\
  \end{bmatrix}
  \text{diag}(\vect{k})
  \vect{\omega}^2 + \vect{d}
\end{align}
with $\vect{G} = \begin{bmatrix}G_p & G_q & G_{a_z}\end{bmatrix}$.
These control effectiveness parameters $\vect{G}$ must be estimated. This can be done by using a LMS-estimator~\cite{smeurAdaptiveIncrementalNonlinear}, or similarly to how in this research the actuator effectiveness factors $k_i$ are estimated, using a Kalman estimator, as explained in the following section. Other stochastic gradient descent methods~\cite{mandicIntrinsicRelationshipLeast2015} could also be considered. This simple model serves as basis of the LOE-detection method.


\section{Methodology} \label{sec:method}

\begin{figure}[thpb]
  \includegraphics[trim=5mm 2mm 0 2mm]{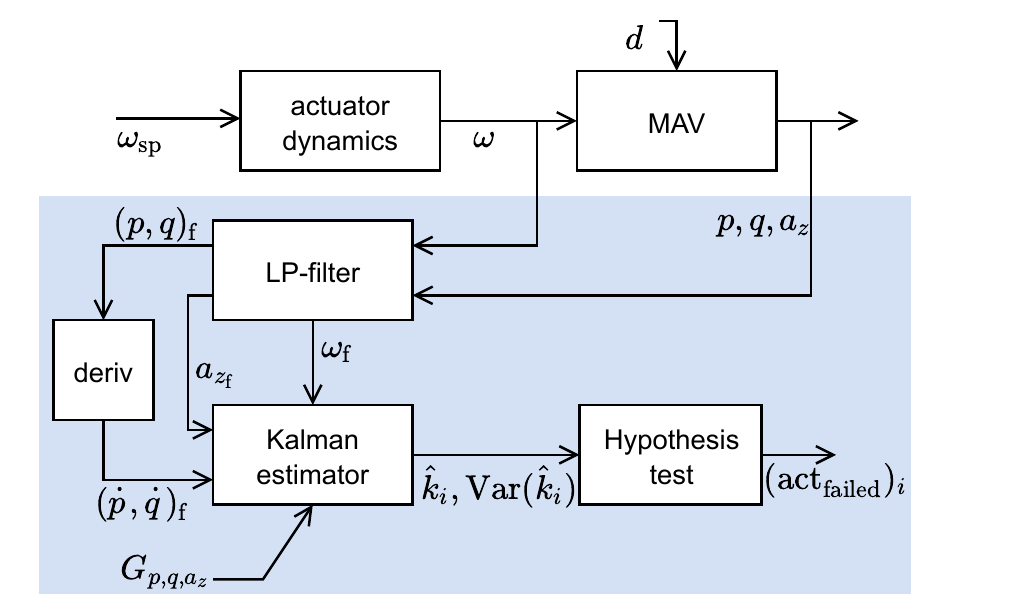}
  \caption{Overview of the loss of effectiveness detection system (blue). The rotor speeds $\omega$ are measured by the ESCs, the rates $p$, $q$, and acceleration $a_z$ are measured by the IMU. Only the rates are differentiated, yet all measurements are filtered with the same bandwidth to keep them synchronized.
  }
  \label{fig:system}
\end{figure}

A block diagram of the loss of effectiveness (LOE) detection system is given in Figure~\ref{fig:system}. The following subsections describe each step, starting with the filtering of the inputs. 
These signals are listed in Table \ref{table:signals}.

\begin{table}[h]
  \centering
  \caption{Signals used for LOE-detection.} \label{table:signals}
  \begin{tabular}{lllll} \toprule
    Signal         & Symbol                          & Unit             & Source       & Contamination \\ \midrule
    Angular rate   & $p$, $q$, $r$                   & rad/s            & IMU          & Noise, bias   \\
    Angular accel. & $\dot{p}$, $\dot{q}$, $\dot{r}$ & rad/$\text{s}^2$ & IMU + filter & Noise, delay  \\
    Proper accel.  & $a_z$                           & m/$\text{s}^2$   & IMU          & Noise, bias   \\
    Rotor speed    & $\omega$                        & rad/s            & ESC          & None          \\
  \end{tabular}
\end{table}

\subsection{Filtering of accelerations and motor speed measurements}



We have to take the derivative of the measured rotational rates to get the angular accelerations. This is a risky operation because of the noise and vibrations. A second order, low-pass filter effectively filters out these disturbances.
The same method as in~\cite{smeurAdaptiveIncrementalNonlinear} is applied, being a filter of the form:
\begin{equation*}
  H(s) = \dfrac{\omega_n^2}{s^2 + 2 \zeta \omega_n s + \omega_n^2}
\end{equation*}
with parameters $\omega_n = 50\ \text{rad/s}$ $\zeta = 0.55$.
Note that by inspecting the raw sensor data --- sampled at 500 Hz --- it was observed that most vibrations are in the $\sim$50-200 Hz range, corresponding to a minimum and maximum rotor speed of the Bebop 2 of $\sim$3000 RPM, resp. $\sim$12000 RPM.
As in~\cite{smeurAdaptiveIncrementalNonlinear}, this same filtering is also applied to the rotor speed $\omega$ and accelerometer $a_z$ measurements in order to keep them synchronized with the gyro measurements.

The angular accelerations are simply approximated by taking the backward difference quotient of the filtered rotational rates:
\begin{equation}
  (\vect{\dot{\Omega}}_\text{f})_k = \dfrac{(\vect{\Omega}_\text{f})_{k} - (\vect{\Omega}_\text{f})_{k-1}}{\Delta{t}}
\end{equation}
this approach is similar to a washout filter~\cite{baconReconfigurableNDIController2001}.

\subsection{Kalman estimator}
We propose a Kalman estimator, its state vector:
\begin{equation}\label{eq:kalman_x}
  \vect{x} = \vect{\hat{k}} =
  \begin{bmatrix}
    \hat{k}_1 & \hat{k}_2 & \hat{k}_3 & \hat{k}_4 \\ 
  \end{bmatrix}^\mathsf{T}
\end{equation}
contains the estimate of the actuator effectiveness scaling factors $k_i$ as defined in Section~\ref{sec:problem}.
The state transition is modeled as a random walk:
\begin{equation*}
  \dot{\vect{x}} = \vect{0}
\end{equation*}

The control effectiveness model \eqref{eq:control_eff} can be rewritten into the following observation model:
\begin{align}\label{eq:obs_model}
  \vect{z}
   & = 
  \vect{H}
  (\vect{\omega})
  \vect{x} \nonumber \\ 
   & =
  \text{diag}(\vect{G})
  \begin{bmatrix}
    1  & -1 & -1 & 1  \\
    1  & 1  & -1 & -1 \\
    -1 & -1 & -1 & -1 \\
  \end{bmatrix}
  \text{diag}(\vect{\omega}^2)
  \vect{x} \nonumber \\ 
   & =
  \begin{bmatrix}
    G_p\w1      & -G_p\w2     & -G_p\w3     & G_p\w4      \\
    G_q\w1      & G_q\w2      & -G_q\w3     & -G_q\w4     \\
    -G_{a_z}\w1 & -G_{a_z}\w2 & -G_{a_z}\w3 & -G_{a_z}\w4 \\
  \end{bmatrix}
  \vect{x}
\end{align}
This model is linear in the state parameters $\vect{x}$. The rotor speeds $\vect{\omega}$ are measured. Its observation vector is:
\begin{equation}\label{eq:kalman_z}
  \vect{z} = 
  \begin{bmatrix}
    \dot{p} & \dot{q} & a_z \\ 
  \end{bmatrix}^\mathsf{T}
\end{equation}
The whole Kalman estimation algorithm is then defined by algorithm \ref{alg:kalman}. 
\begin{algorithm}[h] \label{alg:kalman}
  \Input{$\vect{x}_{k - 1}$, $\vect{P}_{k - 1}$, $\vect{z}_{k}$}
  \BlankLine
  \begin{multicols}{2}
  $\vect{P}_{k|k-1} = \vect{P}_{k - 1} + \vect{Q}$\;
  $\vect{y} = \vect{z}_k - \vect{H}_k \vect{x}_{k - 1}$\;
  $\vect{S} = \vect{R} + \vect{H}_k \vect{P}_{k|k-1} \vect{H}_k^\mathsf{T}$\;
  $\vect{K} = \vect{P}_{k|k-1}\vect{H}_k^\mathsf{T} / \vect{S}$\;
  $\vect{x}_k = \vect{x}_{k - 1} + \vect{K} \vect{y}$\;
  $\vect{P}_{k|k} = (\vect{I} - \vect{K}\vect{H}_k)\vect{P}_{k|k-1}$\;
  \textbf{return} $\vect{x}_k$, $\vect{P}_{k|k}$\;
  \end{multicols}
  \vspace{2mm}
  \caption{\textsc{Kalman estimator}}
\end{algorithm}

To increase the robustness of the algorithm the scaling factors $k_i$ were bound to $\left[0, 1.5\right]$.


\subsection{Hypothesis test}

The estimated scaling factors $\hat{k}_i$ are stochastic variables of which the variances can be found in the diagonal of the covariance matrix:

\begin{equation} \label{eq:kalman_cov}
  \text{diag}(\vect{P}) = \text{Var}({\vect{x}}) = 
  \text{Var}(\{ \hat{k_1}\ \hat{k_2}\  \hat{k_3}\  \hat{k_4}\})
\end{equation}
Now the failure probability per actuator $i$ can be calculated:
\begin{equation*}
  (P_\text{fail})_i = P(\kiest < k_\text{thres})
\end{equation*}
with the estimated scaling factor $\hat{k}_i$, by assuming a parameter $k_\text{thres}$ and by making the same Gaussian distribution assumption as done in designing the Kalman estimator, implying:
\begin{equation}\label{eq:fail_prob}
  P(\kiest < k_\text{thres}) = 1 - \frac{1}{2}\left[1+\operatorname{erf}\left(\frac{k_\text{thres}-\hat{k}_i}{{\sigma^2_k}_i \sqrt{2}}\right)\right]
\end{equation}
with variance ${\sigma^2_k}_i = \text{Var}(\hat{k}_i)$ and $\text{erf}(\cdot)$ being the Gauss error function.

Given the failure probability, a decision can be made on actuator failure:
\begin{equation}
  (\text{act}_\text{failed})_i =
  \begin{cases}
    1 & \text{if}\ (P_\text{fail})_i > (P_\text{fail})_\text{thres} \\
    0 & \text{if}\ (P_\text{fail})_i < (P_\text{fail})_\text{thres} \\
  \end{cases}
\end{equation}
The selection of these thresholds will be discussed in the following section.

\section{Experiments and results} \label{sec:results}

\begin{figure}[thpb]
  \centering
  \includegraphics[scale=1]{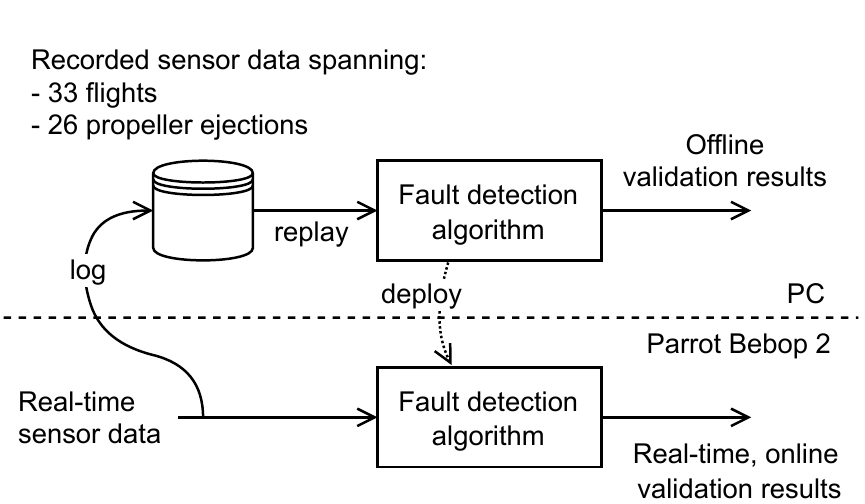}
  \caption{The process used in validating the fault detection algorithm. The exact same algorithm is used in both the offline (PC) environment as in the online real-time environment (Bebop 2). All recorded sensor data is run through the algorithm with 19 different parameter sets.}
  \label{fig:process}
\end{figure}

The LOE-detection algorithm was validated \textit{offline} as well as \textit{online} in real-time as part of an active fault tolerant control system (AFTC). Figure~\ref{fig:process} gives an overview of this process.

The Bebop 2 quadrotor used in the experiments runs PX4~\cite{meierPX4NodebasedMultithreaded2015}. The fault detection algorithm is embedded in PX4 via a C++ library generated with Simulink Coder.

In the following subsections, first of all, a close-up of a single flight is presented, showing the inner workings of the system (\ref{subs:res_single}). Then, the offline replay lets us vary parameters while giving real sensor data as inputs (\ref{subs:res_sensitivity}). We conclude with the key performance metrics (\ref{subs:res_final}) and a brief discussion on ground contact (\ref{subs:res_ground_discussion}).

\subsection{Close up of single flight}\label{subs:res_single}

\begin{figure}[thpb]
  \centering
  \includegraphics[trim=3mm 2mm 0 0]{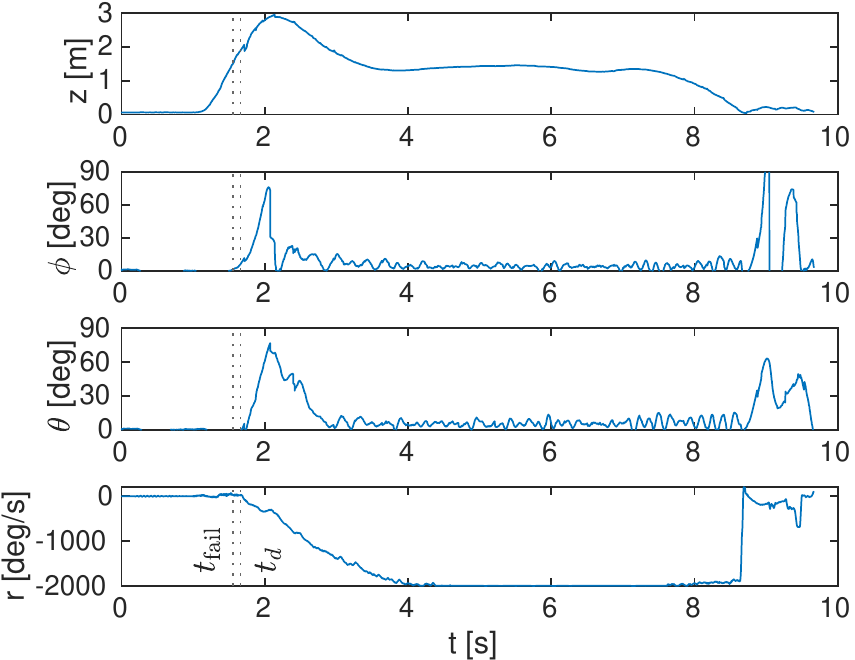}
  \caption{Key states of flight 30. From top to bottom: altitude $z$, roll $\phi$, pitch $\theta$, yaw rate $r$. Actuator 3 fails at $t_\text{fail} = 1.56$, detection is at $t_d = 1.66$. After detection and during increased pitch and roll angles, apogee is reached. Then, the vehicle levels off and reaches a very impressive yaw rate of at least $-2000\ \text{deg/s}$, hitting the angular rate limit of the MPU-6050 gyroscope onboard the Bebop 2 MAV. \label{fig:case_study}}
\end{figure}

\begin{figure}[thpb]
  \centering
  \includegraphics[scale=0.25]{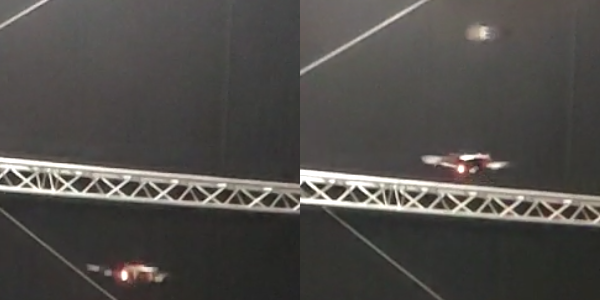}
  \caption{Video frames of the quadrotor at $t_\text{fail}$ (left) and the time of detection $t_d$ (right). The ejected propeller is visible in the right frame, in the top right corner.}
  \label{fig:stills}
\end{figure}

We study flight 30. In this flight the LOE-detector served as input to the AFTC, thus together preventing catastrophic failure. The applied parameters are given in Table \ref{table:params}, Figure~\ref{fig:case_study} shows some of 
the key states of the whole, short, flight. The frames in Figure~\ref{fig:stills} show the moment propeller number three gets ejected and the moment this failure gets detected. 

\begin{figure}[thpb]
  \centering
  \includegraphics[trim=3mm 2mm 0 0]{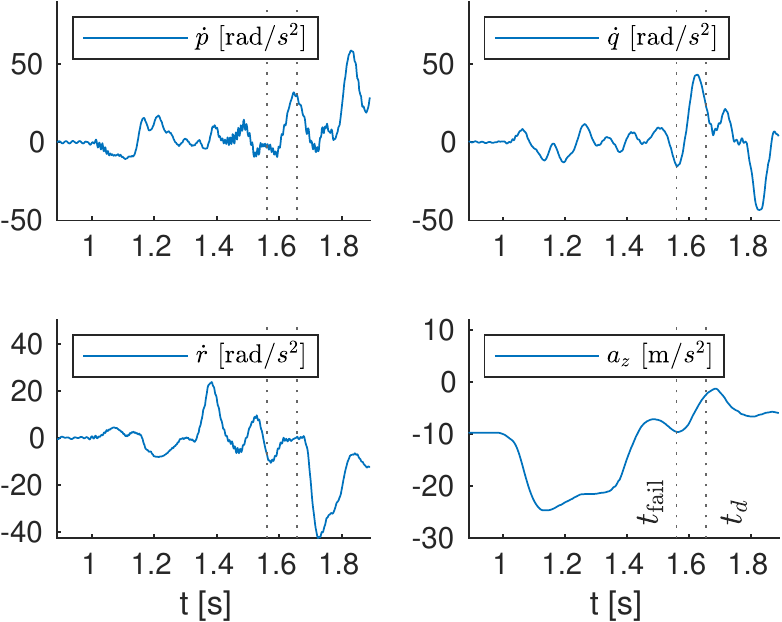}
  \caption{The rotational and linear accelerations, $\dot{p}$, $\dot{q}$ and $a_z$ are the observations to the Kalman estimator. $t_\text{fail}$ and $t_d$ are indicated with the dotted lines. After failure, a positive roll and pitching acceleration can be observed, as well as a reduction in acceleration and an increasingly negative acceleration around the yaw-axis.\label{fig:kalman_z}}
\end{figure}

\begin{figure}[thpb]
  \centering
  \includegraphics[trim=3mm 1mm 0 0]{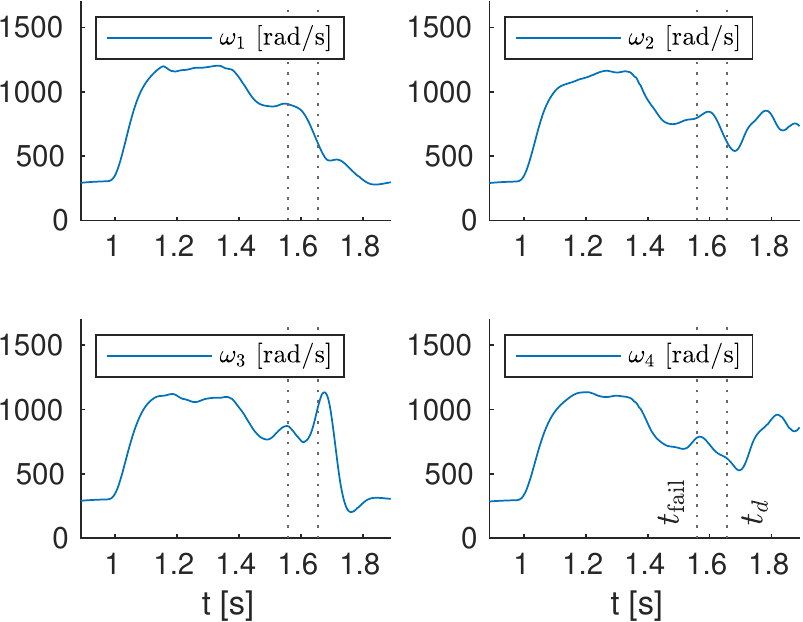}
  \caption{Motor speed measurements during a propeller ejection. In this case, number 3 is the failed one. $t_\text{fail}$ and $t_d$ are indicated with the dotted lines. At detection its setpoint is set to idle by the controller. Actuator 1 is opposite to actuator 3, thus going to idle already \textit{before} detection in an attempt to reach the commanded roll and pitch rates.}\label{fig:rotor_speeds}
\end{figure}

\begin{figure}[thpb]
  \centering
  \includegraphics[trim=3mm 2mm 0 0]{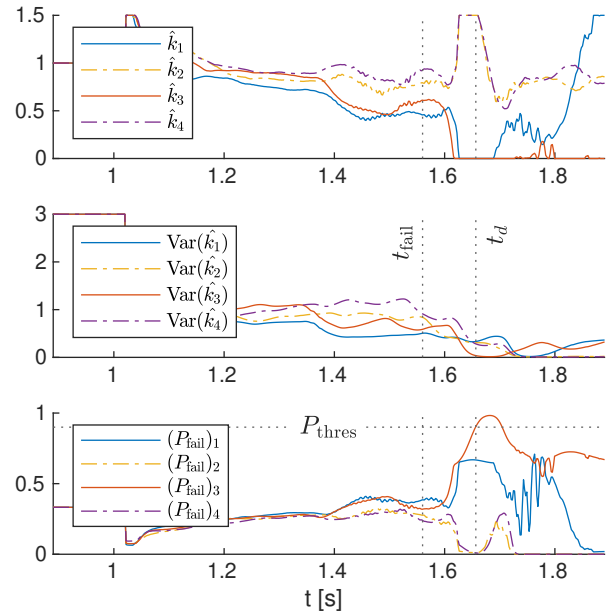}
  \caption{Scaling factors $\kiest$ with variances $\var(\kiest)$ are the Kalman estimator outputs. The failure probabilities $(P_\text{fail})_i = P(\kiest < k_\text{thres})$ with $k_\text{thres} = 0.25$ is the generated signal that triggers the detection. $(P_\text{fail})_\text{thres} = 0.9$ gives the indicated fault detection time $t_d = 1.66\ s$.}
  \label{fig:probs}
\end{figure}

\begin{figure}[thpb]
  \centering
  \includegraphics[trim=3mm 2mm 0 0]{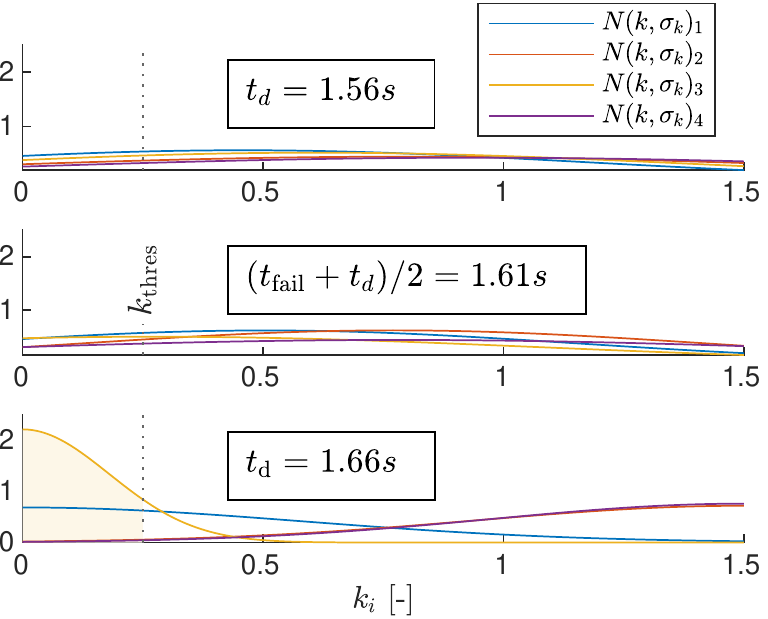}
  \caption{The probability density functions of the actuator effectiveness scaling factors $\kiest$ at $t_\text{fail}$ (top), $(t_\text{fail} + t_d)/2$ (middle) and $t_d$ (bottom). With $k_\text{thres}$ indicated at $k = 0.25$. The shaded area is $(P_\text{fail})_3 = (P_\text{fail})_\text{thres} = 0.9$.}
  \label{fig:prob_density}
\end{figure}

Zooming in on both the failure itself, as well as the detection event, Figure~\ref{fig:kalman_z} shows the accelerations that serve as the observations \ref{eq:kalman_z} to the Kalman estimator. The measured and filtered rotor speeds are shown in Figure~\ref{fig:rotor_speeds}. Then moving to the Kalman estimator itself, in Figure~\ref{fig:probs} the Kalman state vector \eqref{eq:kalman_x}, its variances~\eqref{eq:kalman_cov} and failure probabilities~\eqref{eq:fail_prob} are shown. Figure~\ref{fig:prob_density} shows the probability density function (PDF) of the scaling factors $k_i$ at the indicated timestamps. Following the failure, a clear and distinct shift of the PDF can be observed. Note that when the excitation of the system is low, the variances are trending upward. This is because the observation model~\eqref{eq:obs_model} is an underdetermined system.

\begin{table}[h]
  \centering
  \caption{Used parameters and values.} \label{table:params}
  \vspace{-2mm}
  \begin{tabular}{lll} \toprule
    Parameter                        & Symbol                         & Value                \\ \midrule
    Roll control effectiveness       & $G_p$                          & $100 \times 10^{-6}$ \\
    Pitch control effectiveness      & $G_q$                          & $100 \times 10^{-6}$ \\
    Z-acceleration effectiveness     & $G_{a_z}$                      & $5 \times 10^{-6}$   \\
    Process noise                    & $Q$                            & 0.1                  \\
    Measurement noise                & $R$                            & 1                    \\
    Scaling factor failure threshold & $k_\text{thres}$               & 0.25                 \\
    Failure probability threshold    & $(P_\text{fail})_\text{thres}$ & 0.9                  \\
    Step size                        & $\Delta T$                     & 0.02                 \\
  \end{tabular}
  \vspace{-4mm}
\end{table}



\subsection{Sensitivity analysis across multiple flights}\label{subs:res_sensitivity}

The purpose of this analysis is two-fold: 
1) to investigate the robustness of the output of the algorithm against a variety of flight dynamics and uncertainties that occur during real flights, and 
2) to identify the sensitivity of the parameters.

The sensitivity analysis was done by replaying the algorithm with recorded flight data and varying \textit{one} of the parameters, keeping the others as in Table \ref{table:params}. 
The recorded flight data contains 33 flights, of which 26 experienced a failure per the method described in Section~\ref{sec:problem}. Some of these flights were done inside a wind tunnel experiencing wind speeds up to 10 m/s.

\begin{figure}[thpb]
  \centering
  \includegraphics[trim=3mm 2mm 0 0]{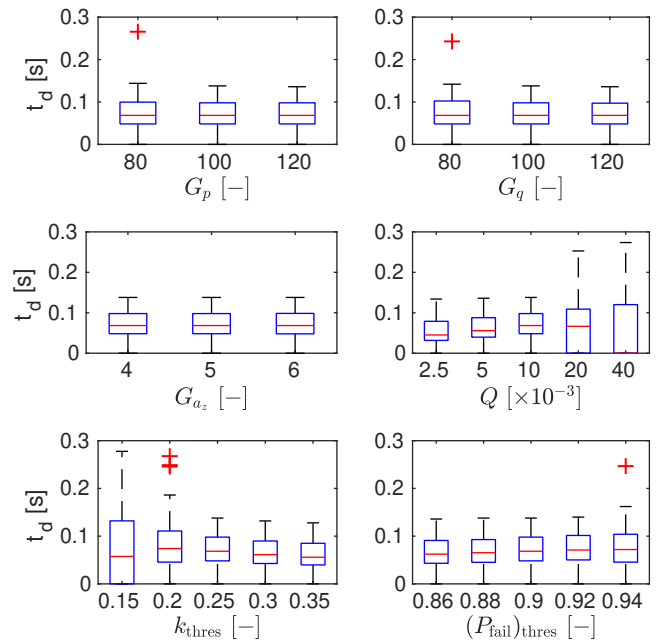}
  \caption{Box plots of the detection delays for varying parameters. The box plots contain the detection delays for the set of 26 flights experiencing an actuator failure. Red crosses indicate the outliers.}
  \label{fig:delay_sens}
\end{figure}

Figure~\ref{fig:delay_sens} presents the detection delays for varying parameters. 
The control effectiveness parameters $\vect{G}$ vary a lot between airframes but can be identified automatically~\cite{smeurAdaptiveIncrementalNonlinear}. They were varied by $\pm 20\%$ and show little sensitivity, therefore they are not considered critical. $Q$, $\pthres$ and $\kthres$ should vary less across airframes, are design variables currently set manually, thus are considered more critical.

Also little sensitivity was observed in the false alarm and missed detection rates against changes in parameter settings. Though, high detection delays $t_d$ tend to correlate with an increasing probability in missed detections.

\subsection{Final results and real-time performance}\label{subs:res_final}

\begin{figure}[thpb]
  \centering
  \includegraphics[trim=3mm 2mm 0 0]{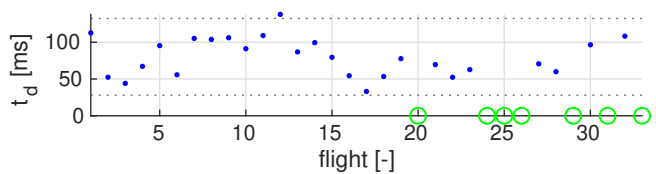}
  \caption{The detection delay $t_d$ of all 26 flights experiencing a propeller ejection. The flights not experiencing a propeller ejection and without any false alarms are indicated as a green circle. The dotted lines indicate the 95\% confidence interval of the detection delays.\label{fig:delays}}
\end{figure}

The following performance metrics are of interest in evaluating fault detection systems~\cite{pattonFAULTTOLERANTCONTROLSYSTEMS}: 1) detection delay, 2) false alarm rate, and 3) missed detection rate.
Note that for our fault scenario --- a sudden failure --- the \textit{detection delay} is of special importance because a quadrotor's lack of redundancy requires fast control reconfiguration to prevent upset conditions.
Figure~\ref{fig:delays} shows the detection delays, no missed detections and no false alarms occurred, with the parameters as in Table \ref{table:params}. 
The 95\% confidence bounds of the detection delay is $\left[ 28, 132 \right]~\text{ms}$.

On the Bebop 2 UAV, equipped with a ARM Cortex-A9 processor, one step of the whole LOE-detection algorithm takes 18 $\muup$s. 
Thus, the total running time is $500~\text{Hz} \cdot 18~\muup\text{s} = 9~\text{ms/sec}$.

\subsection{Discussion on ground contact}\label{subs:res_ground_discussion}
The models in Section \ref{sec:problem} all assume the UAV is in flight. Ground contact will trigger false alarms because the perceived effectiveness of an actuator will be zero. Therefore, we implemented a simple take-off detector that activates the fault detection algorithm after a certain thrust level is reached. Although this works sufficiently in a research setting, more work on land detection is needed to make the system robust against repeated take-offs and landings.

\section{Future work}

Future work should be aimed at maximizing the efficiency and reliability of the method, especially while integrated within the UAV-system:
\begin{itemize}
  \item Investigate possibilities for simplifying the algorithm, ideally reducing the parameter count.
  \item Find better methods for dealing with ground contact.
  \item Evaluate the method under a wider range of flight conditions, properly taking into account the aerodynamic moment occurring during high-speed flight.
\end{itemize}

\addtolength{\textheight}{-10cm}   

\section*{Acknowledgments}

We would like to thank the TU Delft MAVLab for making the drone testing facility available to us, and the PX4-community for building excellent open-source flight control software.


\bibliographystyle{ieeetr}
\bibliography{bibliography}

\end{document}

%% file: diagrams/bebop2_smaller.pdf_tex
\begingroup%
  \makeatletter%
  \providecommand\color[2][]{%
    \errmessage{(Inkscape) Color is used for the text in Inkscape, but the package 'color.sty' is not loaded}%
    \renewcommand\color[2][]{}%
  }%
  \providecommand\transparent[1]{%
    \errmessage{(Inkscape) Transparency is used (non-zero) for the text in Inkscape, but the package 'transparent.sty' is not loaded}%
    \renewcommand\transparent[1]{}%
  }%
  \providecommand\rotatebox[2]{#2}%
  \newcommand*\fsize{\dimexpr\f@size pt\relax}%
  \newcommand*\lineheight[1]{\fontsize{\fsize}{#1\fsize}\selectfont}%
  \ifx\svgwidth\undefined%
    \setlength{\unitlength}{243.77952756bp}%
    \ifx\svgscale\undefined%
      \relax%
    \else%
      \setlength{\unitlength}{\unitlength * \real{\svgscale}}%
    \fi%
  \else%
    \setlength{\unitlength}{\svgwidth}%
  \fi%
  \global\let\svgwidth\undefined%
  \global\let\svgscale\undefined%
  \makeatother%
  \begin{picture}(1,0.81395349)%
    \lineheight{1}%
    \setlength\tabcolsep{0pt}%
    \put(0,0){\includegraphics[width=\unitlength,page=1]{bebop2_smaller.pdf}}%
    \put(0.40605383,0.74832631){\color[rgb]{0,0,0}\makebox(0,0)[lt]{\lineheight{1.25}\smash{\begin{tabular}[t]{l}$\omega_1$\end{tabular}}}}%
    \put(0.93868945,0.74252399){\color[rgb]{0,0,0}\makebox(0,0)[lt]{\lineheight{1.25}\smash{\begin{tabular}[t]{l}$\omega_2$\end{tabular}}}}%
    \put(0.92623345,0.32314049){\color[rgb]{0,0,0}\makebox(0,0)[lt]{\lineheight{1.25}\smash{\begin{tabular}[t]{l}$\omega_3$\end{tabular}}}}%
    \put(0.44019064,0.34549906){\color[rgb]{0,0,0}\makebox(0,0)[rt]{\lineheight{1.25}\smash{\begin{tabular}[t]{r}$\omega_4$\end{tabular}}}}%
    \put(0,0){\includegraphics[width=\unitlength,page=2]{bebop2_smaller.pdf}}%
    \put(0.31793361,0.00483029){\color[rgb]{0,0,0}\makebox(0,0)[rt]{\lineheight{1.25}\smash{\begin{tabular}[t]{r}$z_b$\end{tabular}}}}%
    \put(0.13089779,0.07385289){\color[rgb]{0,0,0}\makebox(0,0)[rt]{\lineheight{1.25}\smash{\begin{tabular}[t]{r}$x_b$\end{tabular}}}}%
    \put(0.10078272,0.2995389){\color[rgb]{0,0,0}\makebox(0,0)[rt]{\lineheight{1.25}\smash{\begin{tabular}[t]{r}$y_b$\end{tabular}}}}%
    \put(0,0){\includegraphics[width=\unitlength,page=3]{bebop2_smaller.pdf}}%
    \put(0.94082219,0.53074375){\color[rgb]{0,0,0}\makebox(0,0)[lt]{\lineheight{1.25}\smash{\begin{tabular}[t]{l}$y_b$\end{tabular}}}}%
    \put(0.69718874,0.78944222){\color[rgb]{0,0,0}\makebox(0,0)[lt]{\lineheight{1.25}\smash{\begin{tabular}[t]{l}$x_b$\end{tabular}}}}%
    \put(0,0){\includegraphics[width=\unitlength,page=4]{bebop2_smaller.pdf}}%
    \put(0.74543958,0.34941143){\color[rgb]{0,0,0}\makebox(0,0)[t]{\lineheight{1.25}\smash{\begin{tabular}[t]{c}$b$\end{tabular}}}}%
    \put(0.88725545,0.4739828){\color[rgb]{0,0,0}\makebox(0,0)[lt]{\lineheight{1.25}\smash{\begin{tabular}[t]{l}$-h$\end{tabular}}}}%
    \put(0,0){\includegraphics[width=\unitlength,page=5]{bebop2_smaller.pdf}}%
    \put(0.19459751,0.07636953){\color[rgb]{0,0,0}\makebox(0,0)[lt]{\lineheight{1.25}\smash{\begin{tabular}[t]{l}$p$\end{tabular}}}}%
    \put(0.18085215,0.2882442){\color[rgb]{0,0,0}\makebox(0,0)[lt]{\lineheight{1.25}\smash{\begin{tabular}[t]{l}$q$\end{tabular}}}}%
    \put(0,0){\includegraphics[width=\unitlength,page=6]{bebop2_smaller.pdf}}%
    \put(0.3530666,0.04605457){\color[rgb]{0,0,0}\makebox(0,0)[lt]{\lineheight{1.25}\smash{\begin{tabular}[t]{l}$r$\end{tabular}}}}%
    \put(0,0){\includegraphics[width=\unitlength,page=7]{bebop2_smaller.pdf}}%
    \put(0.06837127,1.28196139){\color[rgb]{0,0,0}\makebox(0,0)[lt]{\begin{minipage}{1.14055716\unitlength}\raggedright \end{minipage}}}%
    \put(0.08546409,1.33323984){\color[rgb]{0,0,0}\makebox(0,0)[lt]{\begin{minipage}{1.16697338\unitlength}\raggedright \end{minipage}}}%
    \put(-0.04972456,1.3208087){\color[rgb]{0,0,0}\makebox(0,0)[lt]{\begin{minipage}{1.37208708\unitlength}\raggedright \end{minipage}}}%
  \end{picture}%
\endgroup%